# Scope for Machine Learning in Digital Manufacturing

Martin Baumers[1] and Ender Özcan[2]
University of Nottingham

**Abstract:**

**This provocation paper provides an overview of the underlying optimisation problem in the emerging field of Digital Manufacturing. Initially, this paper discusses how the notion of "Digital Manufacturing" is transforming from a term describing a suite of software tools for the integration of production and design functions towards a more general concept incorporating computerised manufacturing and supply chain processes, as well as information collection and utilisation across the product life cycle. On this basis, we use the example of one such manufacturing process, Additive Manufacturing, to identify an integrated multi-objective optimisation problem underlying Digital Manufacturing. Forming an opportunity for a concurrent application of data science and optimisation, a set of challenges arising from this problem is outlined.**

**The emergence of Digital Manufacturing**

In manufacturing, the concept of Digital Manufacturing has arisen and evolved over the recent decades. Initially known as Computer-Integrated Manufacturing, the concept traditionally describes the utilisation of a suite of tools to facilitate the integration of product and process design, with a particular emphasis on jointly optimising "manufacturing before starting the production and supporting ramp-up phases" (Chryssolouris et al., 2009). Cutting across the engineering and operations functions, this collection of digital tools supports process and tooling design, plant layout, advanced visualisation, simulation and concurrent engineering approaches (Slansky, 2008). Broadly, the traditional understanding of Digital Manufacturing can be viewed as part of a change from cost-driven to knowledge-based manufacturing (Westkämper et al., 2007).

Following the emergence and strongly growing relevance of the concept of "Industrie 4.0", the flavour of Digital Manufacturing is changing, however. With a much greater focus on the utilisation of real time data in closed loop control architectures obtained via ubiquitous sensing and computing, the optimisation of networked production facilities has now taken centre stage in Digital Manufacturing. Similarly, the evolved understanding of Digital Manufacturing places a much greater emphasis on flexibility, reconfigurability and resilience in the operation of manufacturing systems (Siemens, 2014). Also described as the application of cyber-physical systems in manufacturing, such systems are based on the idea of creating digital models of processes and products that are expanded throughout various stages in the production flow and later stages in the product life cycle.

**Innovation in underlying manufacturing processes**

However, the emergence of Digital Manufacturing has also resulted in innovation within the underlying manufacturing processes themselves, which remain the centre of manufacturing innovation (Westkämper, 2007). Among other aspects, such process innovation promises to remove existing technological tradeoffs in manufacturing, permit novel supply chain configurations and enable new value propositions in manufacturing.

---

[1] 3D Printing Research Group, Faculty of Engineering, email: martin.baumers@nottingham.ac.uk
[2] Automated Scheduling and Planning Research Group, School of Computer Science, email: ender.ozcan@nottingham.ac.uk

One such process innovation that has received significant attention is Additive Manufacturing (AM), also known as 3D Printing. Forming an archetypal underlying process for Digital Manufacturing, the absence of any physical tooling in AM carries the advantage of being able to deposit complex product geometries without many of the constraints that characterise other manufacturing processes (Baumers et al., 2016). An additional advantage of the additive approach is that very small quantities of products, down to a single unit, can be manufactured efficiently (Tuck et al., 2008).

**An integrated framework**

To build a comprehensive picture of Digital Manufacturing and to begin assessing how Machine Learning techniques can be used to enhance it, it is instructive to take a product life cycle view. Using the example of AM, such an integrated framework can be constructed by combing a Digital Manufacturing model (e.g. Siemens, 2014) with a generic AM process map (Gibson et al., 2010) and a life cycle assessment framework (e.g. BSI, 2012). The resulting model is shown in Figure 1.

As illustrated, the work flow of Digital Manufacturing can thus be understood as an integrated optimisation problem, simultaneously addressing multiple interrelated elements of the Digital Manufacturing work flow, based on information available from and for these steps.

**Setting the scene for Machine Learning and optimisation**

Machine Learning is a growing and diverse field of Artificial Intelligence which studies algorithms that are capable of automatically learning from data and making predictions based on data. Data science techniques, including machine learning have been used in automated manufacturing for a variety of purposes ranging from predictive maintenance, demand forecasting to process monitoring and optimisation (Joseph et al., 2014). With the ever growing amount of data (e.g., from a number of different sensors) in a manufacturing environment, some of the existing issues have turned into Big Data problems requiring an integrated data infrastructure and special knowledge.

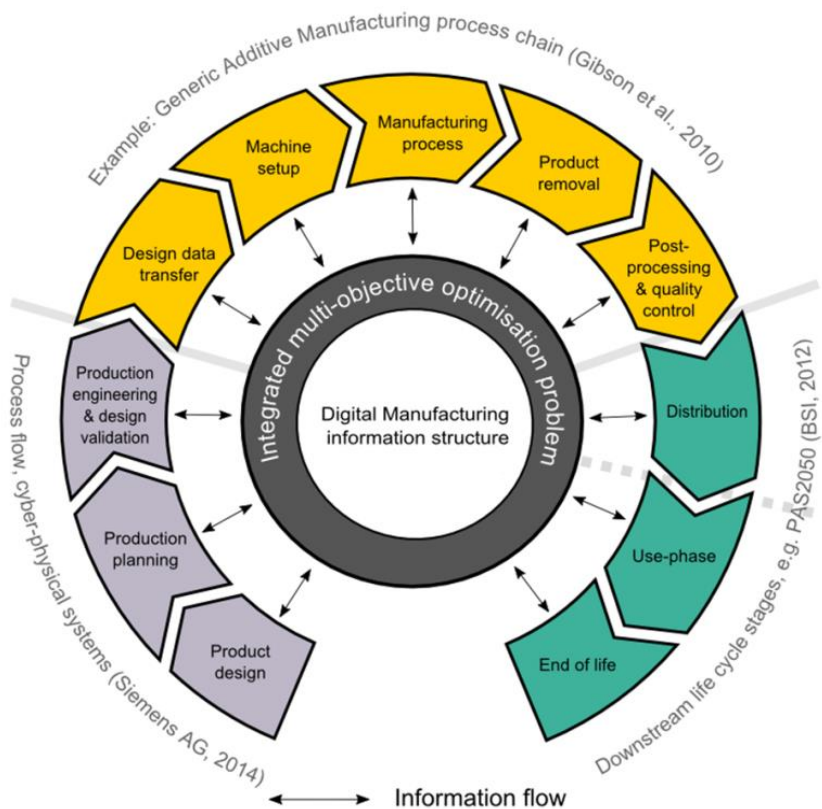

**Figure 1: An integrated framework for the assessment of Digital Manufacturing**

The described integrated framework of Digital Manufacturing indeed embodies a data infrastructure. This provides new opportunities supporting a data-rich integrated environment for concurrent application of data science and optimisation, but gives rise to additional challenges. The framework requires compliance with certain (potentially new) standards as it is extremely desirable for the data science and optimisation techniques (which are not necessarily the same) to communicate directly with each other (Parkes et al., 2015).

Instead of dealing with individual problems which potentially interact and influence each other in a decomposed fashion, this framework describes the conceptual basis for the development of integrated approaches to integrated Digital Manufacturing problems (e.g., scheduling and packing). For the development and application of Machine Learning tools, the challenge among many others faced by industry (Wuest et al., 2016) is to find expertise in the areas of data science and optimisation, considering that each poses its own problems and requires an inherently different skill set.

**Author biographies:**

Dr Martin Baumers is an Assistant Professor of Additive Manufacturing Management at the 3D Printing Research Group in the Faculty of Engineering at the University of Nottingham. Martin is interested in the economic and operational aspects of Digital Manufacturing technologies, with an emphasis on understanding the total cost, build speed and energy consumption of 3D Printing/Additive Manufacturing technologies.

Dr Ender Özcan is a lecturer in Operational Research and Computer Science with the ASAP research group in the School of Computer Science. His research interests and activities lie at the interface of Computer Science, Artificial Intelligence and Operational Research, with a focus on intelligent decision support systems combining data science techniques and (hyper/meta)heuristics applied to real-world problems, such as, timetabling, scheduling, cutting and packing.